\pdfoutput=1
\documentclass[11pt]{article}

\usepackage[]{acl}

\usepackage{times}
\usepackage{latexsym}
\usepackage{amsmath}
\usepackage{amssymb}

\usepackage[T1]{fontenc}
\usepackage[utf8]{inputenc}

\usepackage{microtype}

\usepackage{graphicx,caption}

\usepackage{tabularx}
\usepackage[nobiblatex]{xurl}

\usepackage{colortbl}
\usepackage{enumitem}
\usepackage{adjustbox}
\usepackage{multirow}

\usepackage{soul}

\newcommand{\grayline}{\arrayrulecolor{gray}\hline\arrayrulecolor{black}}

\definecolor{pink1}{RGB}{255,124,124}
\definecolor{pink2}{RGB}{254,220,220}
\definecolor{cyan1}{RGB}{147,201,206}
\definecolor{yellow1}{RGB}{255,254,177}
\definecolor{green1}{RGB}{173,235,173}
\definecolor{violet1}{RGB}{197,175,206}

\DeclareRobustCommand{\hlc}[2]{{\sethlcolor{#1}\hl{#2}}}

\title{\textit{This} Patient Looks Like \textit{That} Patient:\\Prototypical  Networks for Interpretable Diagnosis Prediction\\from Clinical Text}

\author{Betty van Aken$^1$, Jens-Michalis Papaioannou$^1$, Marcel G. Naik$^2$, \\
\textbf{Georgios Eleftheriadis$^2$, Wolfgang Nejdl$^3$, Felix A. Gers$^1$, Alexander Löser}$^1$\vspace{4pt}\\
$^1$ Berliner Hochschule für Technik (BHT), \\
$^2$ Charité Berlin,\\
$^3$ Leibniz University Hannover\\
\fontsize{11}{12}\selectfont \texttt{\{bvanaken,michalis.papaioannou,gers,aloeser\}@bht-berlin.de}, \\
\fontsize{11}{12}\selectfont \texttt{\{marcel.naik,georgios.eleftheriadis\}@charite.de}, 
\texttt{nejdl@L3S.de}}

\begin{document}
\maketitle
\begin{abstract}
The use of deep neural models for diagnosis prediction from clinical text has shown promising results. However, in clinical practice such models must not only be accurate, but provide doctors with interpretable and helpful results. We introduce ProtoPatient, a novel method based on prototypical networks and label-wise attention with both of these abilities. ProtoPatient makes predictions based on parts of the text that are similar to prototypical patients--providing justifications that doctors understand. We evaluate the model on two publicly available clinical datasets and show that it outperforms existing baselines. Quantitative and qualitative evaluations with medical doctors further demonstrate that the model provides valuable explanations for \hbox{clinical decision support.}
\end{abstract}

\section{Introduction}

Medical professionals are faced with a large amount of textual patient information every day. Clinical decision support systems (CDSS) aim to help clinicians in the process of decision-making based on such data. We specifically look at a subtask of CDSS, namely the prediction of clinical diagnosis from patient admission notes. When clinicians approach the task of diagnosis prediction, they usually take similar patients into account (from their own experience, clinic databases or by talking to their colleagues) who presented with typical or atypical signs of a disease. They then compare the patient at hand with these previous encounters and determine the patient’s risk of having the same condition.

In this work, we propose ProtoPatient, a deep neural approach that imitates this reasoning process of clinicians: Our model learns prototypical characteristics of diagnoses from previous patients and bases its prediction for a current patient on the similarity to these prototypes. This results in a model that is both inherently interpretable and provides clinicians with pointers to previous prototypical patients. Our approach is motivated by \citet{chen2019looks} who introduced prototypical part networks (PPNs) for image classification. PPNs learn prototypical parts for image classes and base their classification on the similarity to these prototypical parts. We transfer this work into the text domain and apply it to the extreme multi-label classification task of diagnosis prediction. For this transfer, we apply an additional label-wise attention mechanism that further improves the interpretability of our method by highlighting the most relevant parts of a clinical note regarding a diagnosis.

\begin{figure}[t!]
\centering
  \includegraphics[width=0.49\textwidth]{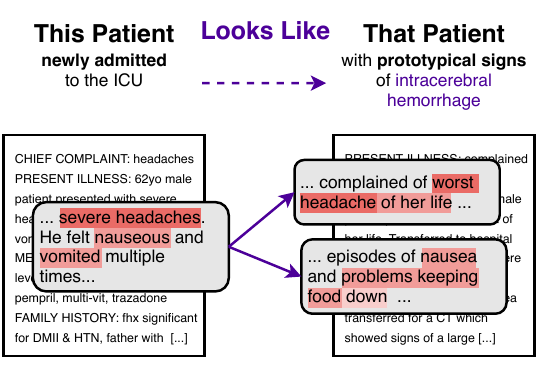}
    \caption{Basic concept of the ProtoPatient method. The model makes predictions for a patient (left side) based on the comparison to prototypical parts of earlier patients (right side).}
\label{fig:intro}
\end{figure}

While deep neural models have been widely applied to outcome prediction tasks in the past \cite{ml-outcomepred},  their black-box nature remains a large obstacle for clinical application \cite{van-aken-etal-2022-see}. We argue that decision support is only possible when model predictions are accompanied by justifications that enable clinicians to follow a lead or to potentially discard predictions. With ProtoPatient we introduce an architecture that allows such decision support. Our evaluation on publicly available data shows that the model can further improve state-of-the-art  performance on predicting clinical outcomes.

\paragraph{Contributions} We summarize the contributions of this work as follows:

\noindent1. We introduce a novel model architecture based on prototypical networks and label-wise attention that enables interpretable diagnosis prediction. The system learns relevant parts in the text and points towards prototypical patients that have led to a certain decision.\\
2. We compare our model against several state-of-the-art baselines and show that it outperforms earlier approaches. Performance gains are especially visible in rare diagnoses.\\
3. We further evaluate the explanations provided by our model. The quantitative results indicate that our model produces explanations that are more faithful to its inner working than post-hoc explanations. A manual analysis conducted by medical doctors further shows the helpfulness of prototypical patients during clinical decision-making.\\
4. We release the code for the model and experiments for reproducibility.\footnote{Public code repository:\\ \url{https://github.com/bvanaken/ProtoPatient}}

\section{Task: Diagnosis Prediction from Admission Notes} 
The task of outcome prediction from admission notes was introduced by \citet{van2021clinical} and assumes the following situation: A new patient $p$ gets admitted to the hospital. Information about the patient is written into an admission note $a_p$. The goal of the decision support system is to identify risk factors in the text and to communicate these risks to the medical professional in charge. For outcome diagnosis prediction in particular, the underlying model determines these risks by predicting the likelihood of a set of diagnoses $C$ being assigned to the patient at discharge. 

\paragraph{Data}
\label{sec:data}
We evaluate our approach on the diagnosis prediction task from the clinical outcome prediction dataset introduced by \citet{van2021clinical}. The data is based on the publicly available MIMIC-III database \cite{johnson2016mimic}. It comprises de-identified data from patients in the Intensive Care Unit (ICU) of the Beth Israel Deaconess Medical Center in Massachusetts in the years 2001-2012. The data includes 48,745 admission notes written in English from 37,320 patients in total. They are split into train/val/test sets with no overlap in patients. The admission notes were created by extracting sections from MIMIC-III discharge summaries which contain information known at admission time such as \textit{Chief Complaint} or \textit{Family History}. The notes are labelled with diagnoses in the form of 3-digit ICD-9 codes that were assigned to the patients at discharge. On average, each patient has 11 assigned diagnoses per admission from a total set of 1266 diagnoses. 

\begin{figure}[t]
  \centering
  \includegraphics[width=74mm]{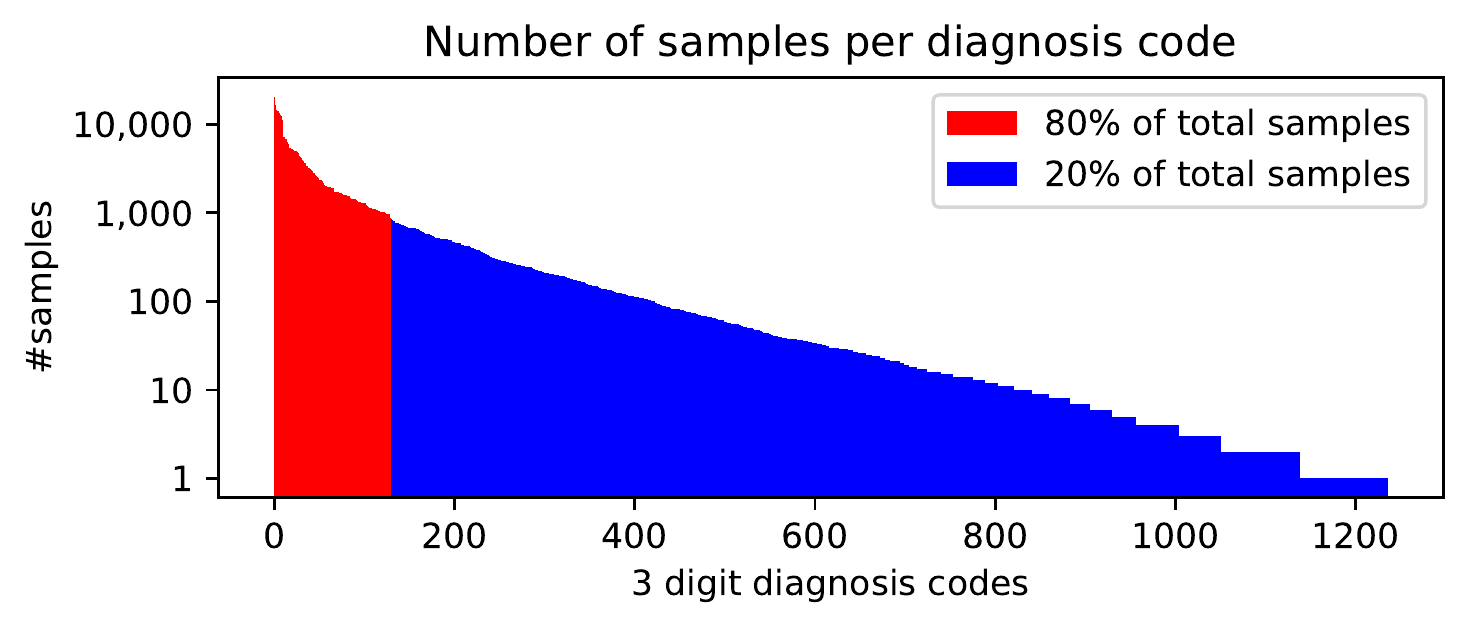}
  \caption{Distribution of ICD-9 diagnosis codes in MIMIC-III training set. }
\label{fig:distribution}
\end{figure}

\begin{figure}[t!]
  \captionsetup{width=1.02\linewidth}
  \includegraphics[width=\columnwidth]{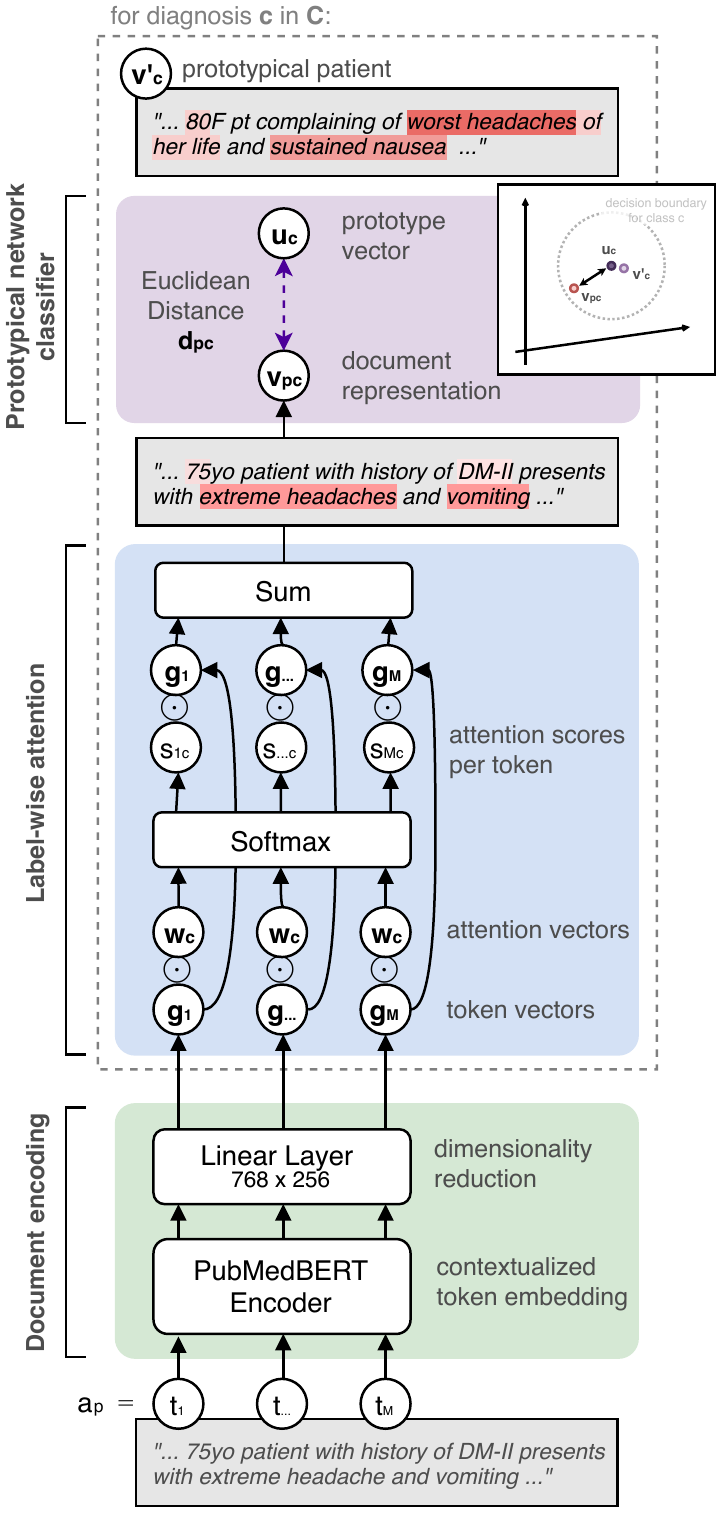}
    \caption{Schematic view of the ProtoPatient method. Starting at the bottom, document tokens get a contextualized encoding and are then transformed into a label-wise document representation $\mathbf{v_{pc}}$. The classifier simply considers the distance of this representation to a learned prototypical vector $\mathbf{u_c}$. The prototypical patient $\mathbf{v'_c}$ is the training example closest to the prototypical vector.}
\label{fig:schema}
\end{figure}

\paragraph{Challenges} Challenges surrounding diagnosis prediction can be divided into two main categories:

\begin{itemize}[leftmargin=2mm]
    \item \textbf{Predicting the correct diagnoses} The number of possible diagnoses is large (>1K) and, as shown in Figure \ref{fig:distribution}, the distribution is extremely skewed. Since many diagnoses only have a few samples, learning plausible patterns is challenging. Further, each admission note describes multiple conditions, some being highly related, while others are not. The text in admission notes is also highly context dependent. Abbreviations like \textit{SBP} (i.a. for \textit{systolic blood pressure} or \textit{spontaneous bacterial peritonitis}) have completely different meanings based on their context. Our models must capture these differences and enable users to check the validity of features used for a prediction.
    
    \item \textbf{Communicating risks to doctors} Apart from assigning scores to diagnoses, for a high-stake task such as diagnosis prediction, a system must be designed for medical professionals to understand and act upon its predictions. Therefore, models must provide faithful explanations for their predictions and give clues that enable further clinical reasoning steps by doctors. These requirements are challenging, since interpretability of models often come with a trade-off in their prediction performance \cite{xai-tradeoff}.
\end{itemize}

\section{Method: ProtoPatient}
To address the challenges above, we propose a novel model architecture called ProtoPatient, which adapts the concept of prototypical networks \cite{chen2019looks} to the extreme multi-label scenario by using label-wise attention and dimensionality reduction. Figure \ref{fig:schema} presents a schematic overview. We further show how our model can be efficiently initialized to improve both speed and performance.

\subsection{Learning Prototypical Representations}
We encode input documents $a_{p}$ ($p$ indexes patients)
into vectors $\mathbf{v_p}$ with dimension $D$ and measure their distance to a learned set of prototype vectors. Each prototype vector $\mathbf{u_c}$ represents a diagnosis $c \in C$ in the dataset. The prototype vectors are learned jointly with the document encoder so that patients with a diagnosis can best be distinguished from patients without it.
As a distance measure we use the Euclidean distance $d_{pc} = ||\mathbf{v_p} -\mathbf{u_c}||_2$ which \citet{snell-prototypical} identified as best suited for prototypical networks. We then calculate the sigmoid $\sigma$ of the negative distances to get a prediction $\hat{y}_{pc} = \sigma{(-d_{pc})}$, so that documents closer to a prototype vector get higher prediction scores. We define the loss $L$ as the binary cross entropy ($BCE$) between $\hat{y}_{pc}$ and the ground truth $y_{pc} \in \{0,1\}$.

\begin{equation}
    L = \sum_p \sum_c BCE(\hat{y}_{pc}, y_{pc})
\end{equation}

\paragraph{Prototype initialization}
\citet{snell-prototypical} define each prototype as the mean of the embedded support set documents. In contrast, we learn the label-wise prototype vectors end-to-end while optimizing the multi-label classification. This leads to better prototype representations, since not all documents are equally representative of a class, as taking the mean would suggest.
However, using the mean of all support documents is a reasonable starting point. 
We set the initial prototype vectors of a class as
$\mathbf{u_{c_{init}}}=\langle \mathbf{v_c} \rangle$, i.e. the mean of all document vectors $\mathbf{v_c}$ with class label $c$ in the training set. We then fine-tune their representation during training. Initial experiments showed that this initialization leads to model convergence in half the number of steps compared to random initialization. 

\paragraph{Contextualized document encoder}
For the encoding of the documents, we choose a Transformer-based model, since Transformers are capable of modelling contextualized token representations. For initializing the document encoder, we use the weights of a pre-trained language model. At the time of our experiments, the PubMedBERT \cite{pubmedbert} model reaches the best results on a range of biomedical NLP tasks \cite{blurb}. We thus initialize our document encoder with PubMedBERT weights\footnote{Model weights from: 
\url{https://huggingface.co/microsoft/BiomedNLP-PubMedBERT-base-uncased-abstract-fulltext}} and further optimize it with a small learning rate during training.

\subsection{Encoding Relevant Document Parts with Label-wise Attention}
Since we face a multi-label problem, having only one joint representation per document tends to produce document vectors located in the center of multiple prototypes in vector space. This way, important features for single diagnoses can get blurred, especially if these diagnoses are rare. To prevent this, we follow the idea of prototypical part networks of selecting parts of the note that are of interest for a certain diagnosis. In contrast to \citet{chen2019looks}, we use an attention-based approach instead of convolutional filters, since attention is an effective way for selecting relevant parts of text. For each diagnosis $c$, we learn an attention vector $\mathbf{w_c}$. To encode a patient note with regard to $c$, we apply a dot product between $\mathbf{w_c}$ and each embedded token $\mathbf{g_{pj}}$, where $j$ is the token index. We then apply a softmax.

\begin{equation}
\label{formula:att1}
    s_{pcj} = softmax(\mathbf{g_{pj}^T \, w_c})
\end{equation}
We use the resulting scores $s_{pcj}$ to create a document representation $\mathbf{v_{pc}}$ as a weighted sum of token vectors.
\begin{equation}
\label{formula:dot}
    \mathbf{v_{pc}} = \sum_j s_{pcj} \, \mathbf{g_{pj}}
\end{equation}

\noindent This way, the document representation for a certain diagnosis is based on the parts that are most relevant to that diagnosis.
We then measure the distance $d_{pc} = ||\mathbf{v_{pc}} -\mathbf{u_c}||_2$ to the prototype vector $\mathbf{u_c}$ based on the diagnosis-specific document representation $\mathbf{v_{pc}}$.

\paragraph{Attention initialization} \label{sec:att-init} The label-wise attention vectors $\mathbf{w_c}$ determine which tokens the final document representation is based on. Therefore, when initializing them randomly, we start our training with document representations which might carry little information about the patient and the corresponding diagnosis. To prevent this cold start, we initialize the attention vectors $\mathbf{w_{c_{init}}}$ with tokens informative to the diagnosis $c$. This way, at training start, these tokens reach higher initial scores $s_{pcj}$. We consider tokens $\tilde{t}$ informative that surpass a TF-IDF threshold of $h$. We then use the average of all embeddings $\mathbf{g_{c\tilde{t}}}$ from $\tilde{t}$ in documents corresponding to the diagnosis.
\begin{equation}
    \mathbf{w_{c_{init}}} = 
    \langle \mathbf{g_{c\tilde{t}}} \rangle
 \end{equation}
 with $\tilde{t} = t : \textrm{tf-idf}(t) > h$.
We found $h$=0.05 suitable to get 5-10 informative tokens per diagnosis.

\subsection{Compressing representations} Label-wise attention vectors for a label space with more than a thousand labels lead to a considerable increase in model parameters and memory load. We compensate this by reducing the dimensionality $D$ of vector representations used in our model. We add a linear layer after the document encoder that both reduces the size of the document embeddings and acts as a regularizer, compressing the information encoded for each document. We find that reducing the dimensionality by one third ($D=256$) leads to improved results compared to the full-size model, indicating that more dense representations are beneficial to our setup.

\subsection{Presenting prototypical patients} For retrieving prototypical patients $\mathbf{v'_c}$ for decision justifications at inference time, we simply take the label-wise attended documents from the training data that are closest to the diagnosis prototype. By presenting their distances to the prototype vector, we can provide further insights about the general variance of diagnosis presentations. Correspondingly, we can also present patients with atypical presentation of a diagnosis by selecting the ones furthest away from the learned prototype.

\section{Evaluating Diagnosis Predictions} \label{sec:experiments}

\subsection{Experimental Setup}

\begin{table*}[t!]
\begin{tabularx}{\textwidth}{lccc}
 & ROC AUC \small{macro} & ROC AUC \small{micro} & PR AUC \small{macro} \\ \hline
HAN \cite{yanghan} & 83.38  \small{$\pm 0.13$} & 96.88  \small{$\pm 0.04$} & 13.56  \small{$\pm 0.01$} \\
HAN + Label Emb \cite{donghlan} & 83.49  \small{$\pm 0.18$} & 96.87  \small{$\pm 0.12$} & 13.07  \small{$\pm 0.14$} \\
HA-GRU \cite{baumelhagru} & 79.94  \small{$\pm 0.57$} & 96.65  \small{$\pm 0.12$} & 9.52  \small{$\pm 1.01$} \\
HA-GRU + Label Emb \cite{donghlan} & 80.54  \small{$\pm 1.67$} & 96.67  \small{$\pm 0.22$} & 10.33  \small{$\pm 1.70$} \\ \grayline
ClinicalBERT \cite{elsentzerclinicalbert} & 80.95  \small{$\pm 0.16$} & 94.54  \small{$\pm 0.93$}  & 11.62  \small{$\pm 0.64$}  \\
DischargeBERT \cite{elsentzerclinicalbert} & 81.17  \small{$\pm 0.30$} & 94.70  \small{$\pm 0.48$} & 11.24  \small{$\pm 0.88$} \\
CORe \cite{van2021clinical} & 81.92  \small{$\pm 0.09$} & 94.00  \small{$\pm 1.10$} & 11.65  \small{$\pm 0.78$} \\
PubMedBERT \cite{pubmedbert} & 83.48  \small{$\pm 0.21$} & 95.47  \small{$\pm 0.22$} & 13.42  \small{$\pm 0.57$} \\ \grayline
Prototypical Network & 81.89  \small{$\pm 0.22$} & 95.23  \small{$\pm 0.01$} & \textcolor{white}{--}9.94  \small{$\pm 0.36$} \\
ProtoPatient & 86.93  \small{$\pm 0.24$} & \textbf{97.32}  \small{$\pm 0.00$} & \textbf{21.16}  \small{$\pm 0.21$} \\
ProtoPatient + Attention Init & \textbf{87.93}  \small{$\pm 0.07$} & 97.24  \small{$\pm 0.02$} & 17.92  \small{$\pm 0.65$} \\ \hline
 &  &  & 
\end{tabularx}
\caption{Results in \% AUC for diagnosis prediction task (1266 labels) based on MIMIC-III data. The ProtoPatient model outperforms the baselines in micro ROC AUC and PR AUC. The attention initialization further improves the macro ROC AUC. $\pm$ values are standard deviations. Label Emb: Label Embeddings. Attention Init: Attention vectors initialized as described in Section \ref{sec:att-init}.}
\label{table:results}
\end{table*}

\paragraph{Baselines} We compare ProtoPatient to hierarchical attention models and to Transformer models  pre-trained on (bio)medical text, representing two state-of-the-arts approaches for ICD coding and outcome prediction tasks, respectively.

\begin{itemize}[leftmargin=*]
    \item \textbf{Hierarchical attention models} Hierarchical Attention Networks (\textbf{HAN}) were introduced by \citet{yanghan}. They are based on bidirectional gated recurrent units, with attention applied on both the sentence and token level. \citet{baumelhagru} built \textbf{HA-GRU} upon this concept using only sentence-wise attention, while adding a label-wise attention scheme comparable to ProtoPatient. \citet{donghlan} further show that pre-initialized \textbf{label embeddings} learned from ICD code co-occurrence improves results for both approaches. We thus evaluate the models with and without label embeddings.\footnote{Note that \citet{donghlan} also propose the H-LAN model, which is a combination of HAN and HA-GRU using label-wise attention on sentence and token level. However, the model is only applicable to smaller label spaces (<100) due to its memory footprint and thus cannot be evaluated on our task.}
    
    \item \textbf{Transformers pre-trained on in-domain text}
    \citet{elsentzerclinicalbert} applied clinical language model fine-tuning on two Transformer models based on the BioBERT model \cite{biobert}. \textbf{ClinicalBERT} was trained on all clinical notes in the MIMIC-III database, and \textbf{DischargeBERT} on all discharge summaries. They belong to the most widely used clinical language models and achieve high scores on multiple clinical NLP tasks. The \textbf{CORe} model \cite{van2021clinical} is also based on BioBERT, but further pre-trained with an objective specific to patient outcomes, which achieved higher scores on clinical outcome prediction tasks. \citet{pubmedbert} introduced \textbf{PubMedBERT} which was, in contrast to the other models, trained from scratch on articles from PubMed Central with a dedicated vocabulary. It is currently the best performing approach on the BLURB \cite{blurb} benchmark.
\end{itemize}

\paragraph{Training}
We train all baselines on the dataset introduced in Section \ref{sec:data}. For training HAN and HA-GRU we use the code and best performing hyperparameters as provided by \citet{donghlan}. We further use their pre-trained ICD-9 label embeddings (for details, see Appendix \ref{sec:label-embeddings}). For training the Transformer-based models and ProtoPatient, we use hyperparameters reported to perform best for BERT-based models by \citet{van2021clinical} and additionally optimize the learning rate and number of warm up steps with a grid search. We further truncate the notes to a context size of 512. See \ref{sec:hyperparameter} for all details on the chosen hyperparameters. We report the scores of all models as an average over three runs with different seeds.

\paragraph{Ablation studies}
ProtoPatient combines three strategies: Prototypical networks, label-wise attention and dimensionality reduction. We conduct ablation studies to measure the impact of each strategy. To this end, we apply both label-wise attention and dimensionality reduction to a PubMedBERT model using a standard classification head. We further train a prototypical network without label-wise attention and ProtoPatient with different dimension sizes. The results are found in Table \ref{table:ablation} and \ref{table:ablation-full}.

\paragraph{Transfer to second data set} Clinical text data varies from clinic to clinic. We want to test whether the patterns learned by the models are transferable to other data sources than MIMIC-III. We use another publicly available dataset from the i2b2 De-identification and Heart Disease Risk Factors Challenge \cite{i2b2} further processed into admission notes by \citet{van2021clinical}. The data consists of 1,118 admission notes labelled with the ICD-9 codes for \textit{chronic ischemic heart disease}, \textit{obesity}, \textit{hypertension}, \textit{hypercholesterolemia} and \textit{diabetes}. We evaluate models without fine-tuning on the new data to simulate a model transfer to another clinic. The resulting scores are reported in Table \ref{table:i2b2}.

\subsection{Results} \label{sec:results} 
\begin{figure}
\centering
  \includegraphics[width=0.42\textwidth]{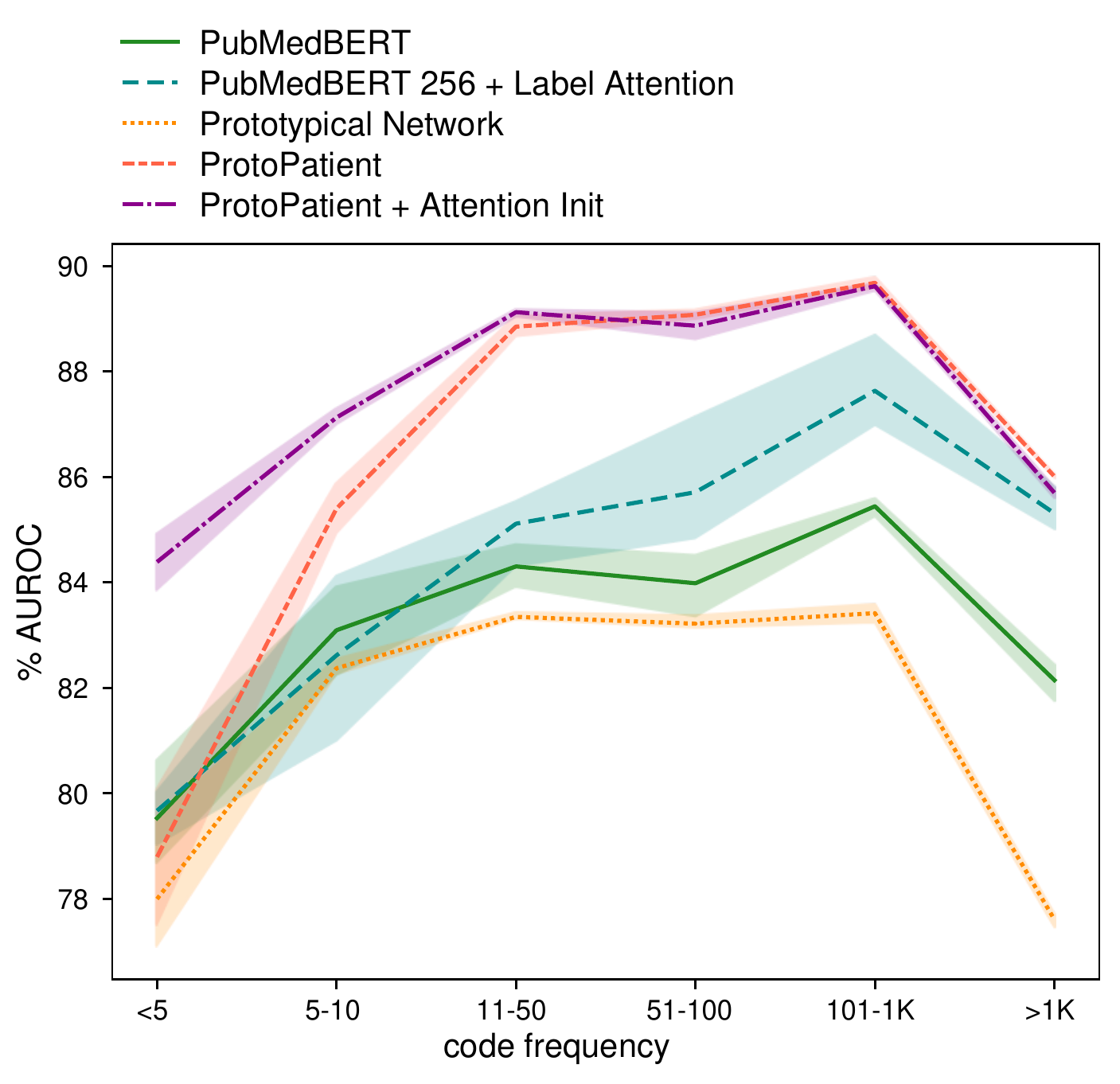}
    \caption{Macro ROC AUC scores regarding the frequency of ICD-9 codes in the training set. ProtoPatient models show the largest performance gain in rare codes ($\leq$100 samples). Attention initialization leads to large improvement for very rare codes ($<$10 samples).}
\label{fig:buckets}
\end{figure}

We present the results of all models on the diagnosis prediction task in Table \ref{table:results}. In addition, we show the macro ROC AUC score across codes depending on their  frequency in the training set in Figure \ref{fig:buckets}. We summarize the main findings as follows.

\paragraph{ProtoPatient outperforms previous approaches} The results show that ProtoPatient achieves the best scores among all evaluated models. Pre-initializing the attention vectors further improves the macro ROC AUC score. Ablation studies show that all components play a role in improving the results. A prototypical network without label-wise attention is not able to capture the extreme multi-label data. PubMedBERT using a standard classification head also benefits from label-wise attention, but not to the same extent. Combining prototypical networks and label-wise attention thus brings additional benefits. The choice of dimension size is another important factor. Using 768 dimensions (the standard BERT base size) appears to lead to over-parameterization in the attention and prototype vectors. Using 256 dimensions also improves generalization, which is shown in producing the best results on the i2b2 data set in Table \ref{table:i2b2}.

\begin{table}
\small
{\renewcommand{\arraystretch}{1.2}%
\setlength\tabcolsep{5pt}
\begin{tabularx}{\columnwidth}{lc}
 & ROC AUC \scriptsize{macro} \\ \hline
\textbf{Dimensionality reduction} & \\
ProtoPatient 768 & 83.56  \small{$\pm 0.17$} \\
ProtoPatient \scriptsize{(our proposed model with $D$=256)} & \textbf{86.93}  \small{$\pm 0.24$} \\
\hline
\textbf{Transformer vs. Prototypical} & \\
PubMedBERT 768 & 83.48  \small{$\pm 0.21$} \\ 
PubMedBERT 768 + Label Attention &  \textbf{84.10}  \small{$\pm 0.25$} \\ 
ProtoPatient 768 & 83.56  \small{$\pm 0.17$} \\ \grayline
\textbf{Label-wise attention} & \\
PubMedBERT 256 & 83.61 \small{$\pm 0.04$} \\ 
PubMedBERT 256 + Label Attention &  \textbf{84.68}  \small{$\pm 0.52$} \\ \hline
\end{tabularx}}
\caption{\textbf{Ablation studies} comparing different dimension sizes and how a standard Transformer (PubMedBERT) performs with additional label-wise attention.}
\label{table:ablation}
\end{table}

\paragraph{Improvements for rare diagnoses}
Figure \ref{fig:buckets} shows that the ROC AUC improvements are particularly large for codes that are rare ($\leq$50 times) in the training set. Prototypical networks are known for their few-shot capabilities \cite{snell-prototypical} which also prove useful in our scenario with mixed label frequencies. For extremely rare codes that appear less than ten times, the attention initialization described in Section \ref{sec:att-init} further improves results. This indicates that the randomly initialized attention vectors need at least a number of samples to learn the most important tokens, and that pre-initializing them can accelerate this process.

\paragraph{PubMedBERT and HAN are the best baselines} The pre-trained PubMedBERT and the HAN model achieve the highest scores among the baselines. Interestingly, PubMedBERT outperforms the Transformer models pre-trained on clinical text. This indicates that training from scratch with a domain-specific vocabulary is beneficial for the task. The scores of the HAN model further emphasize the importance of label-wise attention. The addition of label embeddings to HAN and HA-GRU, however, does not add significant improvements in our case.

\begin{table}[t]
\small
{\renewcommand{\arraystretch}{1.2}%
\begin{tabularx}{\columnwidth}{lc}
 & ROC AUC \small{macro} \\ \hline
PubMedBERT & 82.11 \small{$\pm 0.12$}\\
Prototypical Network & 69.65 \small{$\pm 0.22$} \\
ProtoPatient 768 & 85.28 \small{$\pm  0.49$} \\ 
ProtoPatient & \textbf{87.38} \small{$\pm  0.20$} \\
ProtoPatient + Attention Init & 86.72 \small{$\pm  1.52$} \\ \hline

\end{tabularx}}
\caption{Performance on a second data set based on clinical notes from the \textbf{i2b2 challenge} \cite{i2b2}. ProtoPatient shows the highest degree of transferability. Further metrics shown in Table \ref{table:i2b2-full}.}
\label{table:i2b2}
\end{table}

\section{Evaluating Interpretability}\label{sec:interpret}
We evaluate the interpretability of ProtoPatient with quantitative and qualitative analyses as follows.

\begin{figure}
\centering
  \includegraphics[width=0.45\textwidth]{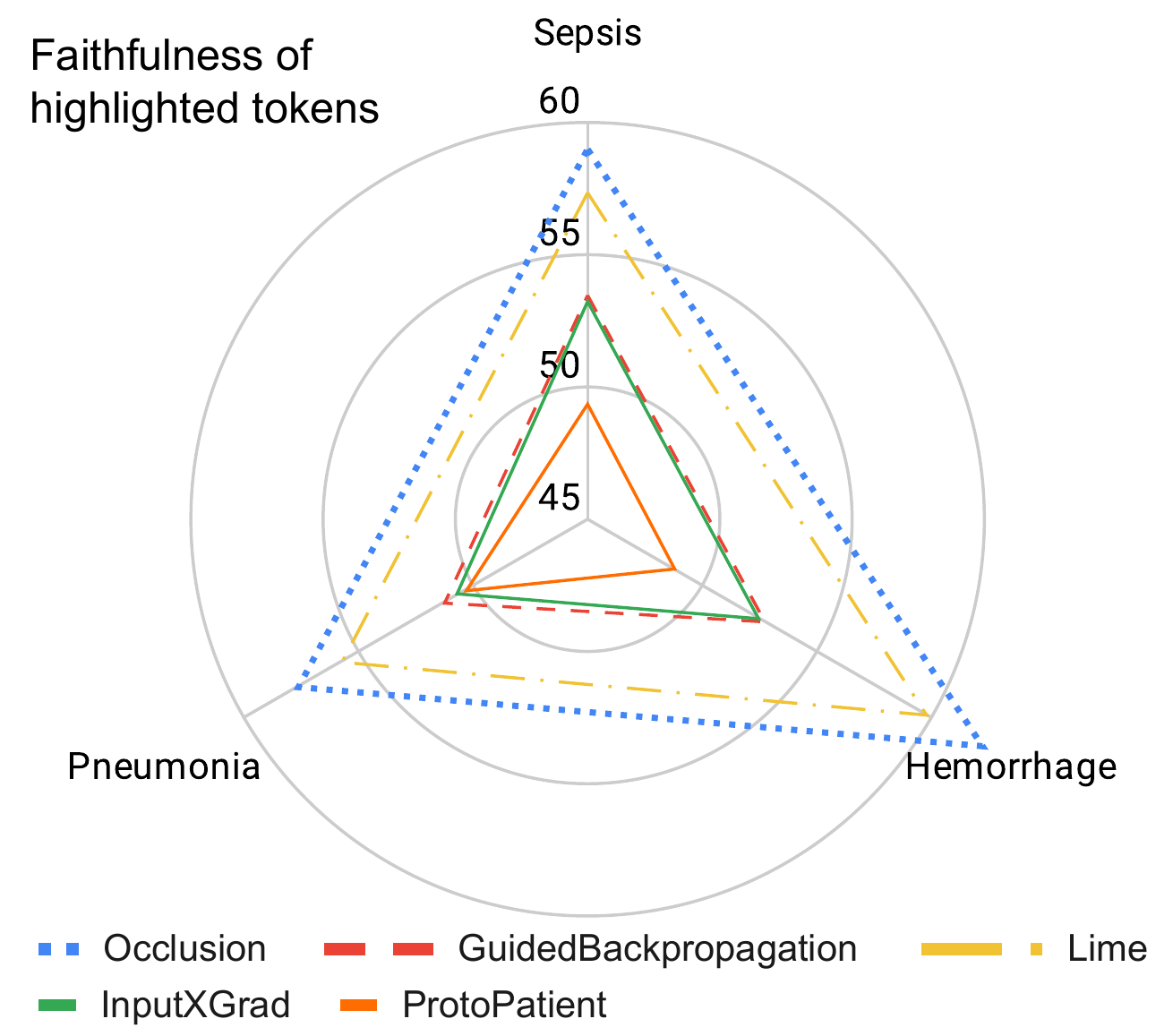}
    \caption{Evaluating faithfulness of highlighted tokens. Lower scores indicate more faithful explanations. ProtoPatient's token highlights are part of the model decision and thus more faithful than post-hoc explanations.}
\label{fig:xai}
\end{figure}

\paragraph{Quantitative study on faithfulness} Faithfulness describes how explanations correspond to the inner workings of a model, a property essential to their usefulness. We apply the explainability benchmark introduced by \citet{xai} to compare the faithfulness of ProtoPatient's token highlights to post-hoc explanation methods. Following the benchmark, faithfulness is measured by incrementally masking highlighted tokens, expecting a steep drop in model performance if the tokens are indeed relevant to the model prediction. See \ref{sec:interpret-full} for details. Due to the high computational costs of the evaluation, we limit our analyses to three diagnoses with a high severity to the ICU: Sepsis, intracerebral hemorrhage and pneumonia. We compare against four common post-hoc explanation methods, namely Lime \cite{lime}, Occlusion \cite{occlusion}, InputXGradient \cite{inputxgradient}, and Gradient Backpropagation \cite{gradientbackpropagation}, which we apply to the PubMedBERT baseline. Figure \ref{fig:xai} shows the results, for which lower scores mean more faithful explanations (i.e. a steeper drop in model performance). We see that ProtoPatient's explanations reach the lowest scores for all three labels, proving that they are more faithful than the post-hoc explanations. This is a result of the interpretable structure of ProtoPatient, in which model decisions are directly based on the highlighted parts. We show these parts, i.e. the tokens that are most frequently highlighted by the model for the three analyzed diagnoses, in \ref{sec:relevant-tokens}.

\begin{table}[b!]
\small
{\renewcommand{\arraystretch}{1.1}%
\begin{tabular}{|llll}
\hline

\multicolumn{4}{|c|}{\begin{tabular}[c]{@{}c@{}}\textbf{Analysis of prototypical patient cases}\\(principal diagnoses)\end{tabular}} \\ \hline

\multicolumn{4}{|c|}{Q1: Prototypical patient shows typical clinical signs} \\ 

 \multicolumn{4}{|c|}{\begin{tabular}{p{2.5cm}|p{2.5cm}}\centering yes & \centering no \end{tabular}} \\
 \hline
  \multicolumn{4}{|c|}{\begin{tabular}{p{2.5cm}|p{2.5cm}}\centering 21 & \centering 2 \end{tabular}} \\

\hline
\multicolumn{4}{|c|}{Q2: Highlighted prototypical parts are relevant} \\ 

 \multicolumn{4}{|c|}{\begin{tabular}{p{2cm}|p{2cm}|p{2cm}}\centering mostly & \centering partially & \centering hardly \end{tabular}} \\ \hline
 
 \multicolumn{4}{|c|}{\begin{tabular}{p{2cm}|p{2cm}|p{2cm}}\centering 21 & \centering 2 & \centering 0 \end{tabular}} \\
\hline

\multicolumn{4}{|c|}{Q3: Prototypical patient is helpful for diagnosis decision} \\ 

 \multicolumn{4}{|c|}{\begin{tabular}{p{2.5cm}|p{2.5cm}}\centering yes & \centering no \end{tabular}} \\ \hline
  \multicolumn{4}{|c|}{\begin{tabular}{p{2.5cm}|p{2.5cm}}\centering 17 & \centering 6 \end{tabular}} \\
\hline

\multicolumn{4}{|c|}{\begin{tabular}[c]{@{}c@{}}\textbf{Analysis of highlighted parts}\\(all diagnoses)\end{tabular}} \\ \hline

\multicolumn{4}{|c|}{\begin{tabular}[c]{@{}c@{}}Q4: Highlighted tokens are relevant for diagnosis\\ (i.e. describe diagnosis, symptoms or risk factors)\end{tabular}} \\ 

\multicolumn{1}{|l|}{} &  \multicolumn{3}{c|}{\begin{tabular}{p{1.6cm}|p{1.6cm}|p{1.6cm}}\centering mostly & \centering partially & \centering hardly \end{tabular}} \\ \hline

\multicolumn{1}{|l|}{TPs} &  \multicolumn{3}{c|}{\begin{tabular}{p{1.6cm}|p{1.6cm}|p{1.6cm}}\centering 78 & \centering 3 & \centering 7 \end{tabular}} \\

\multicolumn{1}{|l|}{FPs} &  \multicolumn{3}{c|}{\begin{tabular}{p{1.6cm}|p{1.6cm}|p{1.6cm}}\centering 50 & \centering 12 & \centering 9 \end{tabular}} \\

\multicolumn{1}{|l|}{FNs} &  \multicolumn{3}{c|}{\begin{tabular}{p{1.6cm}|p{1.6cm}|p{1.6cm}}\centering 22 & \centering 10 & \centering 12 \end{tabular}} \\

\hline

\multicolumn{4}{|c|}{Q5: Important tokens are missing from highlights} \\ 

\multicolumn{1}{|l|}{} &  \multicolumn{3}{c|}{\begin{tabular}{p{2.3cm}|p{2.3cm}}\centering yes & \centering no \end{tabular}} \\ \hline

\multicolumn{1}{|l|}{TPs} &  \multicolumn{3}{c|}{\begin{tabular}{p{2.3cm}|p{2.3cm}}\centering 17 & \centering 71 \end{tabular}} \\

\multicolumn{1}{|l|}{FPs} &  \multicolumn{3}{c|}{\begin{tabular}{p{2.3cm}|p{2.3cm}}\centering 13 & \centering 58 \end{tabular}} \\

\multicolumn{1}{|l|}{FNs} &  \multicolumn{3}{c|}{\begin{tabular}{p{2.3cm}|p{2.3cm}}\centering 2 & \centering 42 \end{tabular}} \\
\hline

\end{tabular}}
\caption{Results of the manual analysis conducted by medical doctors on ProtoPatient outputs. The prototypical patients were analyzed for the principal diagnoses only, while the highlighted parts of the patient letter at hand were analyzed for all diagnoses. Q1..5 denote the questions answered regarding each patient case.}
\label{table:analysis}
\end{table}

\paragraph{Manual analysis by medical doctors} We conduct a manual analysis with two medical doctors (one specialized, one resident) to understand whether highlighted tokens and prototypical patients are helpful for their decisions. They used a demo application of ProtoPatient\footnote{Demo URL available at:\\\url{https://protopatient.demo.datexis.com}} and analyzed 20 random patient letters with 203 diagnoses in total. The results are shown in Table \ref{table:analysis}. The doctors first identified the principal diagnoses and then rated the corresponding prototypical patients presented by the model. Note that some patients have more than one principal diagnosis. In 21 of 23 cases, the prototypical samples were showing typical signs of the respective diagnosis and 17 of them were rated as helpful for making a diagnosis decision. Cases in which they were not helpful included very rare conditions or ones with a strong difference to the specific case. They further analyzed the highlighted tokens for all diagnoses and found that they contained mostly relevant information in 150 cases. Examples of highlighted risk factors judged as plausible were \textit{obesity} known to relate to \textit{diabetes type II}, \textit{untreated hypertension} to \textit{heart failure} or a medication history of \textit{anticoagulant coumadin} to \textit{atrial fibrillation}. They also identified cases in which the highlighted tokens were partially or hardly relevant. In these cases, the highlighted tokens often included stop words or punctuation, indicating that the attention vector failed to learn relevant tokens. This was mainly observed in very frequent diagnoses such as \textit{hypertension} or \textit{anemia}, which corresponds to the lower model performance on these conditions (see Figure \ref{fig:buckets}). This is because conditions very common in the ICU are often either not indicated in the clinical note or not labelled, so that the model cannot learn clear patterns regarding their relevant tokens.

\begin{table*}
\scriptsize
{\renewcommand{\arraystretch}{1.35}%
\begin{tabularx}{\textwidth}{l|X l X}
Admission note & Relevant parts of admission note & similar to & Parts of prototypical patient notes\\ \hline

\multirow{3}{165pt}[-5.5pt]{PRESENT ILLNESS: Patient is a 35-year-old male pedestrian \hlc{pink1}{struck by a bicycle} from behind with \hlc{pink1}{positive loss of consciousness for 6 minutes} at the scene after landing on his head. At arrival at ER \hlc{green1}{patient was confused}, had multiple contusions noted on a \hlc{violet1}{head CT scan including bilateral frontal and right temporal contusions}.  His cervical spine and abdominal examinations were negative radiologically. The patient was then \hlc{cyan1}{transferred to the Emergency Room. Patient had several episodes of vomiting} during flight and during the trauma workup. He was assessed and was \hlc{green1}{intubated for airway protection}. The patient was \hlc{pink1}{given coma score of 9} upon initial assessment. Patient remaining \hlc{violet1}{hemodynamically stable throughout the transfer} and throughout the workup in the ED. […]}

&
\hlc{pink1}{struck by a bicycle} …

\hlc{pink1}{loss of consciousness for 6 minutes} …

\hlc{pink1}{coma score 9} …
& 
\multirow{1}{0pt}[-5pt]{
$\longrightarrow$ }
&
\textbf{cerebral hemorrhage} 

loss of consciousness … 

struck by vehicle … 

with a gcs of 10 …
  \\ \cline{2-4}
& 
\hlc{violet1}{head CT scan} …

\hlc{violet1}{bilateral contusions} …

\hlc{violet1}{hemodynamically stable} …
 & 
\multirow{1}{0pt}[-5pt]{
$\longrightarrow$ }
 &
 \textbf{skull fracture} 
 
 head wound …
 
right and left contusions …

stable blood circulation … 
 \\ \cline{2-4}
& 
\hlc{cyan1}{transferred to Emergency Room} …

\hlc{cyan1}{several episodes of vomiting} …
&
\multirow{1}{0pt}[-3pt]{
$\longrightarrow$ }
&
\textbf{shock} 

patient had multiple episodes of vomiting during the day …
 \\ \cline{2-4}
 
& \hlc{green1}{patient was confused} …

\hlc{green1}{intubated for airway protection} …
&
\multirow{1}{0pt}[-3pt]{
$\longrightarrow$ }
&
\textbf{acute respiratory failure} 

patient was disoriented …

later intubated for protection…
 \\ \hline

\end{tabularx}}
\caption{Exemplary output of ProtoPatient. The model identifies parts in an admission note that are similar to (i.e. \textit{"look like"}) parts from prototypical patient notes seen during training, leading to the prediction of this diagnosis.}
\label{table:example}
\end{table*}

\section{Related Work}

\paragraph{Diagnosis prediction from clinical notes}
Predicting diagnosis risks from clinical text has been studied using different methods. \citet{fakhraie2011s} analyzed the predictive value of clinical notes with bag-of-words and word embeddings. \citet{attention-clinical} experimented with adding attention modules to recurrent neural models. Recently, the use of Transformer models for diagnosis prediction has outperformed earlier approaches. \citet{van2021clinical} applied BERT-based models further pre-trained on clinical cases to predict patient outcomes. However, the black-box nature of these models hinders their application in clinical practice. We therefore introduce ProtoPatient, which uses Transformer representations, but provides interpretable predictions.

\paragraph{Prototypical networks for few-shot learning}
Prototypical networks were first introduced by \citet{snell-prototypical} for the task of few-shot learning. They initialized prototypes as centroids of support samples per episode and applied the approach to image classification tasks. \citet{sun-fewshot-text} adapted the approach to text documents with hierarchical attention layers. Recently, related approaches based on prototypical networks have been used for multiple few-shot text classification tasks \cite{wen21infproc,zhang21kdd,ren20coling,deng20wsdm,feng23compspeech}. In contrast to this body of work, we do not train our model in a few-shot scenario using episodic learning. However, our model shows related capabilities by improving results for diagnoses with few available samples. 

\paragraph{Prototypical networks for interpretable models}
\citet{chen2019looks} used prototypical networks in a different setup to build an interpretable model for image classification. To this end, they learn prototypical parts of images to mimic human reasoning. We adapt their idea and show how to apply it to clinical natural language. 
Recently, \citet{prototext} and \citet{prototex} applied the concept of prototypical networks to text classification and showed how prototypical texts help to interpret predictions. In contrast to their work and following \citet{chen2019looks}, we identify prototypical \textit{parts} rather than whole documents by using label-wise attention. This makes interpreting results easier and enables multi-label classification with over a thousand labels.

\paragraph{Label-wise attention}
\citet{mullenbach-caml} introduced label-wise attention for clinical text with the CAML model. Since then, the method has been further improved by hierarchical attention approaches \cite{baumelhagru,yanghan,donghlan}. Label-wise attention has mainly been used for ICD coding, a task related to diagnosis prediction that differs in the input data: ICD coding is done on notes that describe the whole stay at a clinic. In contrast, outcome diagnosis prediction uses admission notes as input and identifies diagnosis \textit{risks} rather than the diagnoses already mentioned in the text. Our method--combining prototypical networks with label-wise attention--is particularly focused on detecting and highlighting those risks to enable clinical decision support.

\section{Discussion} 

\subsection{Reflection on the Challenges} 

\citet{stopexplaining} urges to stop explaining black-boxes and to build interpretable models instead. With ProtoPatient we introduce a model with a simple decision process--\textit{this patient looks like that patient}--that is understandable to medical professionals and inherently interpretable. An exemplary output is shown in Table \ref{table:example}. 
Our results indicate that the model is able to deal with contextual text in clinical notes, e.g.~when identifying \textit{SBP} as a risk factor for sepsis in \ref{sec:relevant-tokens}. In addition, it improves results on rare diagnoses, which are especially challenging for doctors to detect due to lack of experience and sensitivity towards their signs. Overall, our approach demonstrates that interpretability can be improved without compromising performance. The modularity of the prototype vectors further allows clinicians to modify the model even after training. This can be done by adding prototypes whenever a new condition is found, or by directly defining certain patients as prototypical for the system.

\subsection{Limitations of this work}
 Our model currently learns relations between diagnoses only indirectly, due to the label-wise nature of the classification. However, considering relations or conflicts between diagnoses is an important part of clinical decision-making. One way to include such relations is the addition of a loss term incorporating diagnosis relations, as proposed by \citet{mullenbach-caml}. Another limitation is that the current model only considers one prototype per diagnosis, even though most diagnoses have multiple presentations, varying among patient groups. We therefore propose further research towards including multiple prototypes into the system.
 
\section{Conclusion and Future Work}
In this work, we present ProtoPatient which enables interpretable outcome diagnosis prediction from text. Our approach enhances existing methods in their prediction capability—especially for rare classes—and presents benefits to doctors by highlighting relevant parts in the text and pointing towards prototypical patients. The modularity of prototypical networks can be explored in future research. One promising approach is to introduce multiple prototypes per diagnosis, corresponding to the multiple ways diseases can present in a patient. Prototypes could also be added manually by medical professionals based on patients they consider prototypical. Another approach would be to initialize prototypes from medical literature and compare them to those learned from patients.

\section*{Acknowledgments}
We would like to thank the anonymous reviewers for their valuable feedback. This work was funded by the German Federal Ministry for Economic Affairs and Energy (BMWi) under grant agreements 01MD19003B (PLASS) and 01MK2008MD (Service-Meister), as well as the Federal Ministry of Education and Research (BMBF) under grant agreement 16SV8845 (KIP-SDM).

\bibliography{custom}
\bibliographystyle{acl_natbib}
\newpage
\appendix
\begin{table*}[]
\small
{\renewcommand{\arraystretch}{1.3}%
\begin{tabular}{ll}
Diagnosis \textcolor{white}{--------------} & 15 most attended words - with medical relation to diagnosis \\ \hline

\textbf{Sepsis} & 1. \textbf{hypotension} \textcolor{purple}{symptom}, 2. \textbf{sepsis} \textcolor{orange}{descriptor}, 3. \textbf{fever} \textcolor{purple}{symptom}, 4. \textbf{hypotensive} \textcolor{purple}{symptom}, \\ & 
5. \textbf{fevers} \textcolor{purple}{symptom}, 6. \textbf{septic} \textcolor{orange}{descriptor}, 7. \textbf{lactate} \textcolor{brown}{indicator}, 8. \textbf{shock} \textcolor{orange}{descriptor},  \\ & 9. \textbf{bacteremia} \textcolor{orange}{descriptor}, 10. \textbf{febrile} \textcolor{purple}{symptom}, 11. \textbf{vancomycin} \textcolor{blue}{medication}, 12. \textbf{SBP} \textcolor{teal}{risk factor}, \\ & 13. \textbf{levophed} \textcolor{blue}{medication}, 14. \textbf{swelling} \textcolor{purple}{symptom}, 15. \textbf{cirrhosis} \textcolor{teal}{risk factor} \\ \hline

\textbf{Intracerebral} & 1. \textbf{hemorrhage} \textcolor{orange}{descriptor}, 2. \textbf{bleed} \textcolor{orange}{descriptor}, 3. \textbf{headache} \textcolor{purple}{symptom}, 4. \textbf{ICH} \textcolor{orange}{descriptor}, \\ \textbf{Hemorrhage} & 
5. \textbf{IPH} \textcolor{orange}{descriptor}, 6. \textbf{CT} \textcolor{violet}{diagnostic}, 7. \textbf{weakness} \textcolor{purple}{symptom}, 8. \textbf{stroke} \textcolor{orange}{descriptor}, 9. \textbf{brain} \textcolor{orange}{descriptor}, \\ & 10. \textbf{intracranial} \textcolor{orange}{descriptor}, 11. \textbf{hemorrhagic} \textcolor{orange}{descriptor}, 12. \textbf{intraventricular} \textcolor{orange}{descriptor}, \\ & 13. \textbf{hemorrhages} \textcolor{orange}{descriptor}, 14. \textbf{hemiparesis} \textcolor{purple}{symptom}, 15. \textbf{aphasia} \textcolor{purple}{symptom} \\ \hline

\textbf{Pneumonia} & 1. \textbf{pneumonia} \textcolor{orange}{descriptor}, 2. \textbf{cough} \textcolor{purple}{symptom}, 3. \textbf{PNA} \textcolor{orange}{descriptor}, 4. \textbf{COPD} \textcolor{teal}{risk factor}, \\ & 
5. \textbf{infiltrate} \textcolor{purple}{symptom}, 6. \textbf{distress} \textcolor{red}{complication}, 7. \textbf{fever} \textcolor{purple}{symptom}, 8. \textbf{breath} \textcolor{gray}{\textit{ambiguous}},  \\ & 9. \textbf{hypoxia} \textcolor{purple}{symptom}, 10. \textbf{sputum} \textcolor{purple}{symptom}, 11. \textbf{respiratory} \textcolor{red}{complication}, 12. \textbf{sepsis} \textcolor{red}{complication}, \\ & 13. \textbf{SOB} \textcolor{purple}{symptom}, 14. \textbf{consolidation} \textcolor{purple}{symptom}, 15. \textbf{CAP} \textcolor{orange}{descriptor} \\ \hline

\end{tabular}}
\caption{Words from the test set with the highest attention scores assigned by ProtoPatient. All words are directly related to the diagnoses and mostly describe symptoms or direct descriptors (in various forms). The highlights can therefore help doctors to quickly identify important parts within a note and to compare them to prototypical parts.}
\label{table:tokens}
\end{table*}

\section{Training Details}

\subsection{Label Embeddings for HAN and HA-GRU}
\label{sec:label-embeddings}
We apply label embeddings to the HAN and HA-GRU network as proposed by \citet{donghlan}. In particular, we use the pre-initialized embeddings provided by the authors. Since they use a larger label set, we map their embedding vectors to the ICD-9 groups we use in our study. The mapping is done by averaging all subcodes for one group. If no code is available for an ICD-9 group, we use a randomly initialized vector.

\subsection{Hyperparameter setup}
\label{sec:hyperparameter}
\paragraph{Batch size}
Since we work with 1266 labels, the label-wise attention calculations limit the batch size that fits into memory. We therefore use a batch size of 20 for all models without label-wise attention, 10 for label-wise attention models reduced to a dimensionality of 256 and 5 for the others. Initial experiments showed that the batch sizes have no influence on model performance in our experiments, only on memory consumption and training duration.

\paragraph{Learning rates}
We choose different learning rates for the document encoder weights and the prototype and label-wise attention vectors. Since we expect the encoder weights from the pre-trained Transformer models to be already well aligned with clinical language, we choose a small learning rate between 5e-04 and 5e-06. Since the prototypical diagnosis vectors and the label-wise attention vectors need more adjustments to enable the classification task, we search in a range of 5e-02 and 5e-04. We further apply an AdamW \cite{adamw} optimizer and a linear learning rate scheduler with a warm-up period of 1K to 5K steps. We provide the best hyperparameters per model in the public code repository.

\section{Interpretability Evaluation Details}

\subsection{Measuring faithfulness}\label{sec:interpret-full} We use the evaluation setup proposed by \citet{xai} to measure the faithfulness of ProtoPatient's explanations. The framework evaluates different methods that output saliencies indicating token importance for a model decision. The evaluation then takes place by masking the most salient tokens via multiple thresholds and measuring the model's performance for each one. Thresholds are going from masking only the top 10\% of salient tokens in steps of 10pp until 100\% of tokens are masked. The final faithfulness score is then calculated as the area under the curve of model performance over all thresholds. As a performance measure, we choose macro ROC AUC to stay consistent with the rest of our experiments. We compare tokens highlighted by ProtoPatient's label-wise attention vectors to four post-hoc explanation methods as described in \ref{sec:interpret}. We apply these methods to the PubMedBERT baseline, corresponding to a typical post-hoc explanation approach for an otherwise black-box model.

\subsection{Finding most relevant words per diagnosis}\label{sec:relevant-tokens}
We want to examine which parts of the clinical notes are highlighted by ProtoPatient per diagnosis. To that end, we collect the tokens with the highest attention scores over all training samples per label. We again use the three diagnoses \textit{sepsis}, \textit{intracerebral hemorrhage} and \textit{pneumonia} for a closer analysis. We further map the tokens to their corresponding words. We then let doctors define the words' medical relations to understand which features the model considers important. Table \ref{table:tokens} shows that the most attended words are mainly symptoms or descriptors of the condition at hand, which meets the objective of ProtoPatient to point doctors to relevant parts of a note.

\begin{table*}[t]
\begin{tabularx}{\textwidth}{lccc}
 & ROC AUC \small{macro} & ROC AUC \small{micro} & PR AUC \small{macro} \\ \hline
\textbf{Dimensionality reduction} & & & \\
ProtoPatient 768 & 83.56  \small{$\pm 0.17$} & 96.65  \small{$\pm 0.03$} & 14.36  \small{$\pm 0.16$} \\
ProtoPatient 256 & \textbf{86.93}  \small{$\pm 0.24$} & \textbf{97.32}  \small{$\pm 0.00$} & \textbf{21.16}  \small{$\pm 0.21$} \\ \hline
\textbf{Transformer vs. Prototypical} & & & \\
ProtoPatient 768 & 83.56  \small{$\pm 0.17$} & \textbf{96.65}  \small{$\pm 0.03$} & 14.36  \small{$\pm 0.16$} \\
PubMedBERT 768 + Label Attention & \textbf{84.10}  \small{$\pm 0.25$} & \textbf{96.66}  \small{$\pm 0.17$} & \textbf{19.74}  \small{$\pm 1.27$} \\  \grayline
\textbf{Label-wise attention} & & & \\
PubMedBERT 256 & 83.61 \small{$\pm 0.04$} & 95.76 \small{$\pm 0.05$} & 13.35 \small{$\pm 0.25$} \\ 
PubMedBERT 256 + Label Attention & 84.68  \small{$\pm 0.52$} & 96.86  \small{$\pm 0.14$} & 17.15  \small{$\pm 1.52$} \\ 
ProtoPatient 256 & \textbf{86.93}  \small{$\pm 0.24$} & \textbf{97.32}  \small{$\pm 0.00$} & \textbf{21.16}  \small{$\pm 0.21$} \\ \hline
 &  &  & 
\end{tabularx}
\caption{Full results of our ablation studies. Smaller dimension sizes benefit ProtoPatient, while the effect is less notable on PubMedBERT. Adding label-wise attention, however, increases PubMedBERT results clearly. Overall, the combination of prototypical network, label-wise attention, and reduced dimension in ProtoPatient reaches the best results.}
\label{table:ablation-full}
\end{table*}

\begin{table*}[t]
\begin{tabularx}{\textwidth}{lccc}
 & ROC AUC \small{macro} & ROC AUC \small{micro} & PR AUC \small{macro} \\ \hline
PubMedBERT & 82.11 \small{$\pm 0.12$}& 85.48 \small{$\pm 0.64$}  & 84.38  \small{$\pm 0.54$} \\
PubMedBERT 256 + Label Attention & 79.78 \small{$\pm 5.30$} & 83.43 \small{$\pm 4.54$} & 84.70 \small{$\pm 2.84$} \\
Prototypical Network & 69.65 \small{$\pm 0.22$} & 74.31 \small{$\pm 0.19$} & 78.53 \small{$\pm 0.19$} \\
ProtoPatient 768 & 85.28 \small{$\pm  0.49$} & 88.63 \small{$\pm 0.43$} & 87.78 \small{$\pm 0.10$} \\ 
ProtoPatient & \textbf{87.38} \small{$\pm 0.20$} & \textbf{90.63} \small{$\pm  0.23$} & \textbf{89.72} \small{$\pm 0.24$} \\
ProtoPatient + Attention Init & 86.72 \small{$\pm  1.52$} & 89.84 \small{$\pm  1.16$} & \textbf{89.71} \small{$\pm  1.20$} \\ \hline

 &  &  & 
\end{tabularx}
\caption{Full results of the evaluation on i2b2 data with five classes. Note that the baseline PR AUC is much higher for this task than for the task based on MIMIC-III. ProtoPatient models reach the highest scores, indicating that they are more robust towards changes in text style than the PubMedBERT baselines. The PubMedBERT model with label-wise attention, in particular, shows quite inconsistent results regarding different seeds.}
\label{table:i2b2-full}
\end{table*}

\end{document}